\documentclass[sigconf]{acmart}

\AtBeginDocument{%
  }

\usepackage{multirow}

\usepackage{tikz}
\usetikzlibrary{positioning}
\usetikzlibrary{calc}

\usetikzlibrary{shapes.geometric, arrows, positioning}

\tikzstyle{process} = [rectangle, rounded corners, minimum width=2.5cm, minimum height=1cm, draw=black, fill=gray!10]
\tikzstyle{box} = [rectangle, minimum width=2cm, minimum height=1cm, draw=black]
\tikzstyle{arrow} = [thick,->,>=stealth]
\tikzstyle{textlabel} = [align=center]

\copyrightyear{2025}
\acmYear{2025}
\acmConference[WWW Companion '25]{Companion Proceedings of the ACM Web
Conference 2025}{April 28-May 2, 2025}{Sydney, NSW, Australia}
\acmBooktitle{Companion Proceedings of the ACM Web Conference 2025 (WWW
Companion '25), April 28-May 2, 2025, Sydney, NSW, Australia}
\acmDOI{10.1145/3701716.3715490}
\acmISBN{979-8-4007-1331-6/2025/04}

\acmSubmissionID{srp3572}

\begin{document}

\title{Don't Do RAG:\\When Cache-Augmented Generation is All You Need for Knowledge Tasks}

\author{Brian J Chan}
\authornote{Three authors contributed equally to this research.}
\author{Chao-Ting Chen}\authornotemark[1]
\author{Jui-Hung Cheng}\authornotemark[1]
\affiliation{%
  \institution{Department of Computer Science\\National Chengchi University}
  \city{Taipei}
  \country{Taiwan}
}
\email{{110703065,110703038,110703007}@nccu.edu.tw}

\author{Hen-Hsen Huang}
\affiliation{
  \institution{Insititue of Information Science\\Academia Sinica}
  \city{Taipei}
  \country{Taiwan}}
\email{hhhuang@iis.sinica.edu.tw}


\begin{abstract}
Retrieval-augmented generation (RAG) has gained traction as a powerful approach for enhancing language models by integrating external knowledge sources. However, RAG introduces challenges such as retrieval latency, potential errors in document selection, and increased system complexity. 
With the advent of large language models (LLMs) featuring significantly extended context windows, this paper proposes an alternative paradigm, cache-augmented generation (CAG) that bypasses real-time retrieval. 
Our method involves preloading all relevant resources, especially when the documents or knowledge for retrieval are of a limited and manageable size, into the LLM's extended context and caching its runtime parameters. 
During inference, the model utilizes these preloaded parameters to answer queries without additional retrieval steps.
Comparative analyses reveal that CAG eliminates retrieval latency and minimizes retrieval errors while maintaining context relevance. 
Performance evaluations across multiple benchmarks highlight scenarios where long-context LLMs either outperform or complement traditional RAG pipelines. 
These findings suggest that, for certain applications, particularly those with a constrained knowledge base, CAG provide a streamlined and efficient alternative to RAG, achieving comparable or superior results with reduced complexity.
\end{abstract}






\maketitle

\section{Introduction}
The advent of retrieval-augmented generation (RAG)~\citep{lewis2020retrieval,gao2023retrieval} has significantly enhanced the capabilities of large language models (LLMs) by dynamically integrating external knowledge sources. RAG systems have proven effective in handling open-domain questions and specialized tasks, leveraging retrieval pipelines to provide contextually relevant answers. However, RAG is not without its drawbacks. 
The need for real-time retrieval introduces latency, while errors in selecting or ranking relevant documents can degrade the quality of the generated responses. 
Additionally, integrating retrieval and generation components increases system complexity, necessitating careful tuning and adding to the maintenance overhead.

\begin{figure}
    \centering
\begin{tikzpicture}[scale=0.72, transform shape]
    \def\activecolor{pink!50}
    
    \node[
        draw, box, text width=2cm, minimum height=1.2cm, minimum width=2.5cm, align=center, fill=\activecolor] at (3, 1.6) (ir_model) {\large IR Model};
    
    \node[
        text width=1cm, minimum height=1cm, minimum width=1cm, align=center
    ] at ($(ir_model.center) + (-3.2cm, 0cm)$) (query) {Query};

    \draw[->, line width=0.3mm] (query) to (ir_model);

    \node[
        draw, box, text width=2cm, minimum height=1.2cm, minimum width=2.5cm, align=center, fill=\activecolor
    ] at ($(ir_model.north) + (0cm, 1.2cm)$) (knowledge_source) {\large Knowledge Source};
    
    \draw[<->, line width=0.3mm] (knowledge_source) to (ir_model);

    \node[
        draw, box, minimum height=1.8cm, minimum width=3.6cm, align=center, fill=blue!10
    ] at ($(ir_model.east) + (2.8cm, 0.2cm)$) (LLM_box) {};

    \node at ($(LLM_box.north) + (0cm, -0.3cm)$) (LLM_text) {LLM};

    \node[
        draw, box, text width=1.8cm, minimum height=0.8cm, minimum width=1.8cm, align=center, fill=\activecolor
    ] at ($(LLM_box.south) + (-0.4cm, +0.7cm)$) (knowledge) {Context};

    \draw[->, line width=0.3mm] (ir_model.east) |- (knowledge.west);

    \node[
        draw, box, text width=0.4cm, minimum height=0.8cm, minimum width=0.4cm, align=center, fill=\activecolor
    ] at ($(knowledge.east) + (0.5cm, 0cm)$) (q) {$q$};

    \draw[->, line width=0.3mm] 
    ($(query.east) + (1.2cm, 0cm) $) -| ($(query.south) + (1.2cm, -0.6cm)$) -| (q.south);

    \node[
        draw, box, text width=0.4cm, minimum height=0.9cm, minimum width=0.4cm, align=center, fill=\activecolor
    ] at ($(LLM_box.east) + (1.2cm, 0cm)$) (response) {$r$};

    \draw[->, line width=0.3mm] (LLM_box.east) to (response);

    \node at ($(LLM_box.south) + (-1cm, -1cm)$) {\large Retrieval-Augmented Generation};
    
    \draw[line width=0.3mm, dashed] (-0.8cm, -0.7cm) to (10.5cm, -0.7cm);


    \node[
        draw, box, text width=2cm, minimum height=1.2cm, minimum width=2.8cm, align=center
    ] at (2.2, -2) (knowledge_source) {\large Knowledge Source};

    \node[
        draw, box, minimum height=1.8cm, minimum width=3.6cm, align=center, fill=blue!10
    ] at ($(LLM_box.south) + (0cm, -3.8cm)$) (LLM_box2) {};
    
    \node[
        text width=1cm, minimum height=1cm, minimum width=1cm, align=center
    ] at ($(LLM_box2.center) + (-7.3cm, 0cm)$) (query) {Query};

    \node at ($(LLM_box2.north) + (0cm, -0.3cm)$) (LLM_text) {LLM};

    \node[
        draw, box, text width=1.8cm, minimum height=0.9cm, minimum width=1.8cm, align=center, fill=white
    ] at ($(LLM_box2.south) + (-0.4cm, +0.7cm)$) (knowledge_cache) {Context Cache};

    \draw[->, line width=0.3mm, dashed] (knowledge_source.south)
        |- node[text width=3cm, pos=0.75, below] {\large Offline Preloading}
        (knowledge_cache);

    \node[
        draw, box, text width=0.4cm, minimum height=0.9cm, minimum width=0.4cm, align=center, fill=\activecolor
    ] at ($(knowledge_cache.east) + (0.5cm, 0cm)$) (q) {$q$};

    \draw[->, line width=0.3mm] 
    (query.south) -| ($(query.south) + (0cm, -0.8cm)$) -| (q.south);

    \node[
        draw, box, text width=0.4cm, minimum height=0.9cm, minimum width=0.4cm, align=center, 
        fill=\activecolor
    ] at ($(LLM_box2.east) + (1.2cm, 0cm)$) (response) {$r$};

    \draw[->, line width=0.3mm] (LLM_box2.east) to (response);
    
    \node at ($(LLM_box2.south) + (-1cm, -1cm)$) {\large Cache-Augmented Generation};

    \node[draw, fill=\activecolor, minimum width=0.1cm, minimum height=0.1cm] (legend1) at (6.7cm, 3.9cm) {};
    \node[anchor=west] at ($(legend1) + (0.15cm, 0cm)$) {\large Active during inference};

\end{tikzpicture}
\caption{Comparison of Retrieval-Augmented Generation (RAG) and our Cache-Augmented Generation (CAG) Workflows: The pink-shaded components represent the processes active during real-time inference. In RAG (top section), the IR model retrieves relevant information from the knowledge source, and both the retrieved knowledge and query are processed by the LLM during inference, introducing retrieval latency. In contrast, CAG (bottom section) preloads and caches knowledge offline, allowing the LLM to process only the query during inference, eliminating retrieval overhead and ensuring a more efficient generation process.}\label{fig:overview}
\Description{This figure compares the RAG and CAG workflows, highlighting the differences in retrieval and cache usage during inference.}
\end{figure}
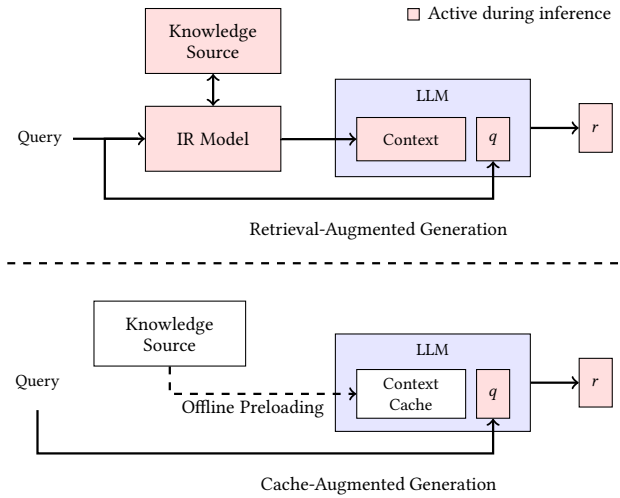

This paper proposes an alternative paradigm, cache-augmented generation (CAG), leveraging the capabilities of long-context LLMs to address these challenges. 
Instead of relying on a retrieval pipeline, as shown in Figure~\ref{fig:overview}, our approach involves preloading the LLM with all relevant documents in advance and precomputing the key-value (KV) cache~\citep{pope2023efficiently}, which encapsulates the inference state of the LLM. 
The preloaded context enables the model to provide rich, contextually accurate answers without the need for additional retrieval during runtime. 
This approach eliminates retrieval latency, mitigates retrieval errors, and simplifies system architecture, all while maintaining high-quality responses by ensuring the model processes all relevant context holistically.

Recent advances in long-context LLMs have extended their ability to process and reason over substantial textual inputs. 
For example, Llama 3.1~\citep{dubey2024llama} was trained with a 128K context length, and its effective context length is 32K in Llama 3.1 8B and 64K in Llama 3.1 70B~\citep{hsieh2024ruler}.
This 32K to 64K context window is sufficient for storing knowledge sources such as internal company documentation, FAQs, customer support logs, and domain-specific databases, making it practical for many real-world applications.
By accommodating larger context windows, these models can assimilate extensive information in a single inference step, making them well-suited for tasks like document comprehension, multi-turn dialogue, and summarization of lengthy texts. 
This capability eliminates the dependency on real-time retrieval, as all necessary information can be preloaded into the model. 
These developments create opportunities to streamline workflows for knowledge-intensive tasks, potentially reducing or even eliminating the need for traditional RAG systems.

Recent studies~\citep{leng2024long,li-etal-2024-retrieval} have investigated the performance of long-context models in RAG tasks, revealing that state-of-the-art models like GPT-o1, GPT-4, and Claude 3.5 can effectively process large amounts of retrieved data, outperforming traditional systems in many scenarios. 
Findings suggest that as long as all documents fit within the extended context length, traditional RAG systems can be replaced by these long-context models. 
Similarly, \citet{lu2024turboragacceleratingretrievalaugmentedgeneration} has demonstrated the benefits of precomputed KV caching to improve efficiency, albeit with the need for position ID rearrangement to enable proper functioning. 
Nonetheless, these methods remain vulnerable to retrieval failures inherent to RAG systems.

Through a series of experiments comparing traditional RAG workflows with our proposed approach, we identify scenarios where long-context LLMs outperform RAG in both efficiency and accuracy. 
By addressing the technical and practical implications, this paper aims to provide insights into when and why CAG may serve as a streamlined, effective alternative to RAG, particularly for cases where the documents or knowledge for retrieval are of limited, manageable size. 
Our findings challenge the default reliance on RAG for knowledge integration tasks, offering a simplified, robust solution to harness the growing capabilities of long-context LLMs.
Our contributions are threefold as follows: 
\begin{itemize}
    \item \textbf{Efficient Alternative to RAG}: We introduced a novel approach leveraging long-context LLMs with preloaded documents and precomputed KV caches, mitigating retrieval latency, errors, and system complexity.
    \item \textbf{Quantitative Analysis}: We conducted extensive experiments showing scenarios where long-context LLMs outperform traditional RAG systems, especially with manageable knowledge bases.
    \item \textbf{Practical Insights}: This work provided actionable insights into optimizing knowledge-intensive workflows, demonstrating the viability of retrieval-free methods for specific applications.
    Our CAG framework is released publicly.\footnote{\url{https://github.com/hhhuang/CAG}} 
\end{itemize}

\section{Methodology}
Our CAG framework leverages the extended context capabilities of long-context LLMs to enable retrieval-free knowledge integration. 
By preloading external knowledge sources, such as a collection of documents \(\mathcal{D} = \{d_1, d_2, \dots\}\), and precomputing the KV cache \(\mathcal{C}_{\text{KV}}\), we address the computational challenges and inefficiencies inherent to real-time retrieval in traditional RAG systems. 
The operation of our framework is divided into three phases:  
\begin{enumerate}
\item \textbf{External Knowledge Preloading}\\
In this phase, a curated collection of documents \(\mathcal{D}\) relevant to the target application is preprocessed and formatted to fit within the model's extended context window. The LLM \(\mathcal{M}\), with parameters \(\theta\), processes \(\mathcal{D}\), transforming it into a precomputed KV cache:  
   \begin{equation}
   \mathcal{C}_{\text{KV}} = \texttt{KV-Encode}(\mathcal{D})    
   \end{equation}

   This KV cache, which encapsulates the inference state of the LLM, is stored on disk or in memory for future use. The computational cost of processing \(\mathcal{D}\) is incurred only once, regardless of the number of subsequent queries.

\item \textbf{Inference}\\
During inference, the precomputed KV cache \(\mathcal{C}_{\text{KV}}\) is loaded alongside the user's query $q$. 
The LLM utilizes this cached context to generate responses:  
   \begin{equation}\label{eq:inference}
   r = \mathcal{M}(\mathcal{D} \oplus q) = \mathcal{M}(q \mid \mathcal{C}_{\text{KV}})    
   \end{equation}
   By preloading the external knowledge, this phase eliminates retrieval latency and reduces risks of errors or omissions that arise from dynamic retrieval. 
   The combined prompt $\mathcal{D} \oplus q$ ensures a unified understanding of both the external knowledge and the user query.

\item \textbf{Cache Reset} \\
To maintain system performance across multiple inference sessions, the KV cache, stored in memory, can be reset efficiently. As the KV cache grows in an append-only manner with new tokens $q = \left(t_1, t_2, \dots, t_k \right)$ sequentially appended, resetting involves truncating these new tokens. 
This allows for rapid reinitialization without reloading the entire cache from disk, ensuring sustained speed and responsiveness.
\end{enumerate}

The proposed methodology offers several significant advantages over traditional RAG systems:
\begin{itemize}
    \item \textbf{Reduced Inference Time}: 
    By eliminating the need for real-time retrieval, the inference process becomes faster and more efficient, enabling quicker responses to user queries.  
    \item \textbf{Unified Context}: 
    Preloading the entire knowledge collection into the LLM provides a holistic and coherent understanding of the documents, resulting in improved response quality and consistency across a wide range of tasks.  
    \item \textbf{Simplified Architecture}: 
    By removing the need to integrate retrievers and generators, the system becomes more streamlined, reducing complexity, improving maintainability, and lowering development overhead.
\end{itemize}

Looking forward, our approach is poised to become even more powerful with the anticipated advancements in LLMs. 
As future models continue to expand their context length, they will be able to process increasingly larger knowledge collections in a single inference step. 
Additionally, the improved ability of these models to extract and utilize relevant information from long contexts will further enhance their performance. 
These two trends will significantly extend the usability of our approach, enabling it to handle more complex and diverse applications. 
Consequently, our methodology is well-positioned to become a robust and versatile solution for knowledge-intensive tasks, leveraging the growing capabilities of next-generation LLMs.

\section{Experiments}
\subsection{Experimental Setup}
To evaluate the effectiveness of our proposed method, we conducted experiments using two widely recognized question-answering benchmarks: the Stanford Question Answering Dataset (SQuAD) 1.0 ~\citep{rajpurkar-etal-2016-squad} and the HotPotQA dataset ~\citep{yang2018hotpotqa}.
These datasets provide complementary challenges, with SQuAD focusing on precise, context-aware answers within single passages and HotPotQA emphasizing multi-hop reasoning across multiple documents.  
Each of both datasets consists of documents \(\mathcal{D} = \{d_1, d_2, \dots\}\) paired with questions \(\mathcal{Q} = \{q_1, q_2, \dots\}\) and golden responses \(\mathcal{R} = \{r_1, r_2, \dots\}\). 
These datasets provide a robust platform for assessing both single-context comprehension and complex multi-hop reasoning. 

To investigate how different levels of reference text length impact retrieval difficulty, we created three test sets for each dataset, varying the size of the reference text. 
For example, in the HotPotQA-small configuration, we sampled 16 documents \(\mathcal{D}_s \subset \mathcal{D}\) from the HotPotQA document set to form a long reference text. 
QA pairs associated with \(\mathcal{D}_s\) were selected as test instances. 
The same methodology was applied to create test sets for SQuAD. 

The dataset statistics are summarized in Table~\ref{tab:datasets}. 
As the number of documents (and hence the length of the reference text) increases, the task becomes more challenging, particularly for RAG systems. 
Longer reference texts increase the difficulty of accurately retrieving the correct information, which is crucial for LLMs to generate high-quality responses.

\begin{table}[tbhp]
    \centering
    \caption{The SQuAD and HotPotQA test sets with varying reference text lengths, highlighting the number of documents, questions, and associated responses for each configuration.}
    \label{tab:datasets}
    \begin{tabular}{llccr}
        \toprule
        Source & Size & \# Docs & \# Tokens & \# QA Pairs  \\
        \midrule
        \multirow{3}{*}{HotPotQA}
          & Small & 16 & 21k & 1,392 \\
          & Medium & 32 & 43k & 1,056 \\
          & Large & 64 & 85k & 1,344 \\
        \midrule
        \multirow{3}{*}{SQuAD}
         & Small & 3 & 21k & 500 \\
         & Medium & 4 & 32k & 500 \\
         & Large & 7 & 50k & 500 \\
        \bottomrule
    \end{tabular}
\end{table}

The primary task involves generating accurate and contextually relevant answers \(\mathcal{\hat{R}} = \{\hat{r}_1, \hat{r}_2, \dots \}\) for the SQuAD and HotPotQA questions, based on the respective preloaded passages. 
By leveraging the precomputed key-value cache \(\mathcal{C}_{\text{KV}} = \texttt{KV-Encode}(\mathcal{D})\), our system generates responses \(\hat{r}_i = \mathcal{M}(q_i \mid \mathcal{C}_{\text{KV}})\) without relying on retrieval mechanisms during inference. 
This unified approach allows for direct performance comparisons against traditional RAG systems, highlighting the strengths and limitations of our method across diverse QA challenges. 

The experiments were executed on Tesla V100 32G $\times$ 8 GPUs. 
For all experiments, we used Llama 3.1 8B~\citep{dubey2024llama} as the underlying LLM across all systems, including both the RAG baselines and our proposed method. 
This model supports input sizes of up to 128k tokens, enabling the processing of extensive contexts. 
For our proposed method, the context of each dataset was preloaded into the model via a precomputed KV cache. 
For SQuAD, the documents \(\mathcal{D}_{\text{S}}\) were encoded into a KV cache \(\mathcal{C}_{\text{KV}}^{\text{S}} = \texttt{KV-Encode}(\mathcal{D}_{\text{S}})\).
For HotPotQA, similarly, the documents \(\mathcal{D}_{\text{H}}\) were encoded into \(\mathcal{C}_{\text{KV}}^{\text{H}} = \texttt{KV-Encode}(\mathcal{D}_{\text{H}})\).  
These caches were stored offline and loaded during inference to eliminate the need for real-time retrieval, ensuring comprehensive access to all relevant information for each dataset.

Our experiments were conducted on both the SQuAD and HotPotQA datasets to evaluate the performance of different systems in terms of similarity to ground-truth answers, measured using BERTScore~\citep{zhangbertscore}. 
Each dataset—SQuAD and HotPotQA—was evaluated separately, with retrieval systems configured to fetch passages exclusively from the respective dataset to ensure focused and fair evaluation.

\subsection{Baseline Systems}

The baseline RAG systems were implemented using the LlamaIndex framework,\footnote{\url{https://www.llamaindex.ai/framework}} employing two retrieval strategies: BM25 for sparse retrieval and OpenAI Indexes for dense retrieval. 
The details of each baseline system are as follows: 

\begin{enumerate}
    \item \textbf{Sparse Retrieval System (BM25)}: 
   The first baseline system employed BM25 indexes for retrieval. 
   BM25, a sparse retrieval algorithm, ranks documents based on term frequency-inverse document frequency (TF-IDF) and document length normalization. 
   Given a query \( q_i \), BM25 retrieves the top-\(k\) passages \(\mathcal{P}_k = \{p_1, p_2, \dots, p_k\}\) from the indexed collection \(\mathcal{D}\).
   These passages were then passed to the generator, \(\mathcal{M}\), to synthesize answers:  
   \begin{equation}\label{eq:rag}
   \hat{r}_i = \mathcal{M}(q_i \mid \mathcal{P}_k)    
   \end{equation}
   BM25 provides a robust and interpretable retrieval mechanism, suited for tasks involving keyword matching.
    \item \textbf{Dense Retrieval System (OpenAI Indexes)}: 
   The second baseline utilized OpenAI indexes,\footnote{\url{https://cookbook.openai.com/examples/evaluation/evaluate_rag_with_llamaindex}} which employ dense embeddings to represent both documents and queries in a shared semantic space. For a query \( q_i \), dense retrieval selects the top-\(k\) passages \(\mathcal{P}_k\) that semantically align with the query, offering improved contextual understanding compared to sparse methods. 
   These passages were similarly passed to the generator for answer synthesis as Equation~\ref{eq:rag}.
   This system is particularly effective for questions requiring nuanced contextual matching beyond exact term overlap.
\end{enumerate}

For the RAG baselines, the top-1, top-3, top-5, and top-10 retrieved passages were used for inference. 
In contrast, our CAG utilized the preloaded context specific to each dataset to generate answers without retrieval constraints. 


\subsection{Results}

\begin{table}
  \caption{Experimental Results}
  \label{tab:results}
  \begin{tabular}{llccc}
    \toprule
    & & & HotPotQA & SQuAD \\
    Size & System & Top-$k$ & BERT-Score & BERT-Score \\
    \midrule
    \multirow{11}{*}{Small}
      & \multirow{4}{*}{Sparse RAG}\
       & 1 & 0.6788 & 0.7214 \\ 
      & & 3 & 0.7626 & 0.7616 \\ 
      & & 5 & 0.7676 & 0.7608 \\ 
      & & 10 & 0.7521 & 0.7584 \\ 
      \cline{2-5}
      & \multirow{4}{*}{Dense RAG}\
       & 1 & 0.7164 & 0.6216 \\ 
      & & 3 & 0.7582 & 0.7106 \\ 
      & & 5 & 0.7481 & 0.7334 \\ 
      & & 10 & 0.7576 & 0.7586 \\ 
      \cline{2-5}
    & CAG (Ours) & & \textbf{0.7951} & \textbf{0.7695} \\ 
    \midrule
    \multirow{11}{*}{Medium}
      & \multirow{4}{*}{Sparse RAG}\
       & 1 & 0.6592 & 0.6902 \\ 
      & & 3 & 0.7546 & 0.7301 \\ 
      & & 5 & 0.7633 & 0.7298 \\ 
      & & 10 & 0.7458 & 0.7262 \\ 
      \cline{2-5}
      & \multirow{4}{*}{Dense RAG}\
       & 1 & 0.6973 & 0.5871 \\ 
      & & 3 & 0.7432 & 0.6702 \\ 
      & & 5 & 0.7322 & 0.6890 \\ 
      & & 10 & 0.7308 & 0.7310 \\ 
      \cline{2-5}
    & CAG (Ours) & & \textbf{0.7821} & \textbf{0.7383} \\ 
    \midrule
    \multirow{11}{*}{Large}
      & \multirow{4}{*}{Sparse RAG}\
       & 1 & 0.6616 & 0.7254 \\ 
      & & 3 & 0.7463 & 0.7634 \\ 
      & & 5 & \textbf{0.7535} & 0.7658 \\ 
      & & 10 & 0.7345 & 0.7613 \\ 
      \cline{2-5}
      & \multirow{4}{*}{Dense RAG}\
       & 1 & 0.7020 & 0.6070 \\ 
      & & 3 & 0.7409 & 0.7018 \\ 
      & & 5 & 0.7234 & 0.7286 \\ 
      & & 10 & 0.7374 & 0.7590 \\ 
      \cline{2-5}
    & CAG (Ours) & & 0.7407 & \textbf{0.7734} \\ 
    \bottomrule
  \end{tabular}
\end{table}

\begin{table}
    \caption{Response Time (Seconds) Comparison on HotPotQA}
    \label{tab:generation-time}
    \begin{tabular}{llcr}
    \toprule
    Size & System & Retrieval & Generation \\
    \midrule
    \multirow{6}{*}{Small}
        & Sparse RAG, Top-3   & 0.0008 & 0.7406 \\
        & Sparse RAG, Top-10  & 0.0012 & 1.5595 \\
        & Dense RAG, Top-3  & 0.4849 & 1.0093 \\
        & Dense RAG, Top-10 & 0.3803 & 2.6608 \\
        & CAG          & -      & 0.8512 \\
        & In-Context Learning      & -      & 9.3197 \\
    \midrule
    \multirow{6}{*}{Medium}
        & Sparse RAG, Top-3   & 0.0008 & 0.7148  \\
        & Sparse RAG, Top-10  & 0.0012 & 1.5306  \\
        & Dense RAG, Top-3  & 0.4140 & 0.9566 \\
        & Dense RAG, Top-10 & 0.4171 & 2.6361 \\
        & CAG          & -      & 1.4078 \\
        & In-Context Learning      & -      & 26.3717 \\
    \midrule 
    \multirow{6}{*}{Large}
        & Sparse RAG, Top-3   & 0.0008 & 0.6667 \\
        & Sparse RAG, Top-10  & 0.0012 & 1.5175 \\
        & Dense RAG, Top-3  & 0.4123 & 0.9331 \\
        & Dense RAG, Top-10 & 0.4100 & 2.6447 \\
        & CAG          & -      & 2.2631 \\
        & In-Context Learning      & -      & 92.0824 \\
    \bottomrule
    \end{tabular}
\end{table}

As shown in Table~\ref{tab:results}, the experimental results highlight key distinctions between our proposed CAG approach and RAG systems. 
CAG consistently achieved the highest BERTScore in most cases, outperforming both sparse and dense RAG methods. 
By preloading the entire reference text from the test set, our method is immune to retrieval errors, ensuring holistic reasoning over all relevant information. 
This advantage is particularly evident in scenarios where RAG systems struggle with retrieving incomplete or irrelevant passages, leading to suboptimal answer generation.

However, as the data size increases, the performance gap between CAG and RAG narrows slightly, aligning with prior findings that long-context LLMs may experience degradation when handling very long contexts~\citep{li2024longcontextllmsstrugglelong}. 
Additionally, the fact that sparse RAG outperforms dense RAG suggests that the datasets may not be sufficiently challenging, allowing traditional sparse retrieval to effectively capture most relevant information without requiring deeper semantic retrieval.
Despite these factors, the results underscore the robustness and efficiency of CAG, particularly for tasks that require a unified understanding of the source material. 
By fully leveraging the long-context capabilities of Llama 3.1, our approach bypasses retrieval challenges and maintains superior performance in retrieval-free knowledge integration.

\usetikzlibrary{patterns, positioning}
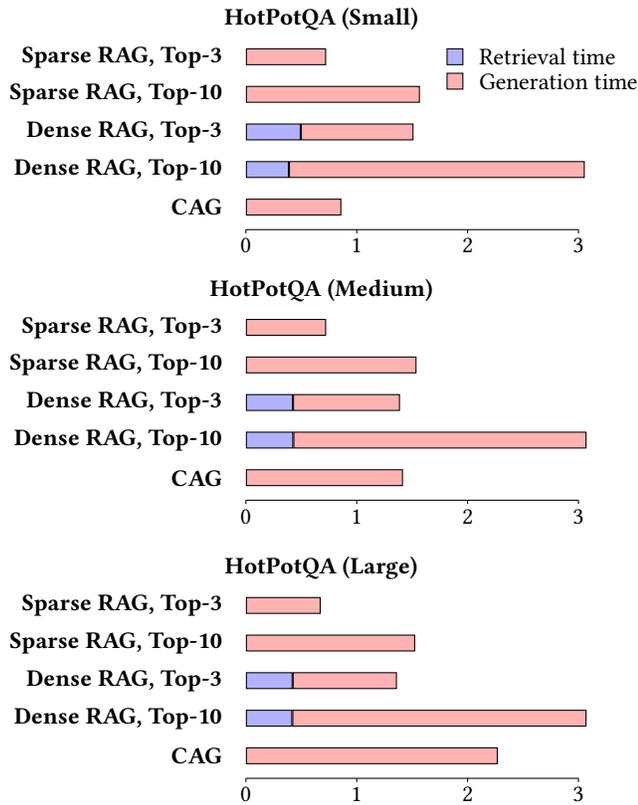
\begin{figure}
\begin{tikzpicture}
    \tikzstyle{bar} = [draw, minimum width=0.05cm, minimum height=0.1cm, anchor=west]
    \def\fullwidth{4.5cm}
    \def\ir-color{blue!30}
    \def\gen-color{red!30}
    \tikzstyle{label}=[anchor=east, minimum height=0.1cm, minimum width=2cm, text width=3cm]

    \draw[-] (0, -2.3) -- (4.4214, -2.3) node[below] {\textbf{}};
    \foreach \x in {0, 1.4738, 2.9476, 4.4214} {
        \draw (\x, -2.3) -- (\x, -2.35);
        \pgfmathsetmacro\dividedX{\x*3.0547/4.5}
        \node[below] at (\x, -2.3) {\pgfmathprintnumber[fixed,precision=0]{\dividedX}};
    }

    \draw[-] (0, -5.9) -- (4.4214, -5.9) node[below] {\textbf{}};
    \foreach \x in {0, 1.4738, 2.9476, 4.4214} {
        \draw (\x, -5.9) -- (\x, -5.95);
        \pgfmathsetmacro\dividedX{\x*3.0547/4.5}
        \node[below] at (\x, -5.9) {\pgfmathprintnumber[fixed,precision=0]{\dividedX}};
    }

    \draw[-] (0, -9.6) -- (4.4214, -9.6) node[below] {\textbf{}};
    \foreach \x in {0, 1.4738, 2.9476, 4.4214} {
        \draw (\x, -9.6) -- (\x, -9.65);
        \pgfmathsetmacro\dividedX{\x*3.0547/4.5}
        \node[below] at (\x, -9.6) {\pgfmathprintnumber[fixed,precision=0]{\dividedX}};
    }
    
    \node (sm-label) at (1, 0.5) {\textbf{HotPotQA (Small)}};
    \node[anchor=east] at (-0.2, 0) {\textbf{Sparse RAG, Top-3}};
    \node[anchor=east] at (-0.2, -0.5) {\textbf{Sparse RAG, Top-10}};
    \node[anchor=east] at (-0.2, -1) {\textbf{Dense RAG, Top-3}};
    \node[anchor=east] at (-0.2, -1.5) {\textbf{Dense RAG, Top-10}};
    \node[anchor=east] at (-0.2, -2) {\textbf{CAG}}; 

    \node at (1, -3.1) {\textbf{HotPotQA (Medium)}};
    \node[anchor=east] at (-0.2, -3.6) {\textbf{Sparse RAG, Top-3}};
    \node[anchor=east] at (-0.2, -4.1) {\textbf{Sparse RAG, Top-10}};
    \node[anchor=east] at (-0.2, -4.6) {\textbf{Dense RAG, Top-3}};
    \node[anchor=east] at (-0.2, -5.1) {\textbf{Dense RAG, Top-10}};
    \node[anchor=east] at (-0.2, -5.6) {\textbf{CAG}}; 

    \node at (1, -6.8) {\textbf{HotPotQA (Large)}};
    \node[anchor=east] at (-0.2, -7.3) {\textbf{Sparse RAG, Top-3}};
    \node[anchor=east] at (-0.2, -7.8) {\textbf{Sparse RAG, Top-10}};
    \node[anchor=east] at (-0.2, -8.3) {\textbf{Dense RAG, Top-3}};
    \node[anchor=east] at (-0.2, -8.8) {\textbf{Dense RAG, Top-10}};
    \node[anchor=east] at (-0.2, -9.3) {\textbf{CAG}}; 
    \node[bar, minimum width=0.234*\fullwidth, fill=\gen-color] (sm-bm25-top3-gen) at (0, 0) {};
    \node[bar, minimum width=0.511*\fullwidth, fill=\gen-color] (sm-bm25-top10-gen) at (0, -0.5) {};
    \node[bar, minimum width=0.159*\fullwidth, fill=\ir-color] (sm-dense-top3-ir) at (0, -1) {};
    \node[bar, minimum width=0.33*\fullwidth, fill=\gen-color, right=0mm of sm-dense-top3-ir] (sm-dense-top3-gen) {};
    \node[bar, minimum width=0.124*\fullwidth, fill=\ir-color] (sm-dense-top10-ir) at (0, -1.5) {};
    \node[bar, minimum width=0.871*\fullwidth, fill=\gen-color, right=0mm of sm-dense-top10-ir] (sm-dense-top10-gen) {};
    \node[bar, minimum width=0.279*\fullwidth, fill=\gen-color] (sm-cag-gen) at (0, -2) {};
    
    \node[bar, minimum width=0.234*\fullwidth, fill=\gen-color] (md-bm25-top3-gen) at (0, -3.6) {};
    \node[bar, minimum width=0.501*\fullwidth, fill=\gen-color] (md-bm25-top10-gen) at (0, -4.1) {};
    \node[bar, minimum width=0.136*\fullwidth, fill=\ir-color] (md-dense-top3-ir) at (0, -4.6) {};
    \node[bar, minimum width=0.313*\fullwidth, fill=\gen-color, right=0mm of md-dense-top3-ir] (md-dense-top3-gen) {};
    \node[bar, minimum width=0.137*\fullwidth, fill=\ir-color] (md-dense-top10-ir) at (0, -5.1) {};
    \node[bar, minimum width=0.863*\fullwidth, fill=\gen-color, right=0mm of md-dense-top10-ir] (md-dense-top10-gen) {};
    \node[bar, minimum width=0.461*\fullwidth, fill=\gen-color] (md-cag-gen) at (0, -5.6) {};

    \node[bar, minimum width=0.218*\fullwidth, fill=\gen-color] (lg-bm25-top3-gen) at (0, -7.3) {};
    \node[bar, minimum width=0.497*\fullwidth, fill=\gen-color] (lg-bm25-top10-gen) at (0, -7.8) {};
    \node[bar, minimum width=0.135*\fullwidth, fill=\ir-color] (lg-dense-top3-ir) at (0, -8.3) {};
    \node[bar, minimum width=0.305*\fullwidth, fill=\gen-color, right=0mm of lg-dense-top3-ir] (lg-dense-top3-gen) {};
    \node[bar, minimum width=0.134*\fullwidth, fill=\ir-color] (lg-dense-top10-ir) at (0, -8.8) {};
    \node[bar, minimum width=0.866*\fullwidth, fill=\gen-color, right=0mm of lg-dense-top10-ir] (lg-dense-top10-gen) {};
    \node[bar, minimum width=0.741*\fullwidth, fill=\gen-color] (lg-cag-gen) at (0, -9.3) {};

    \node[draw, fill=\ir-color, minimum width=0.1cm, minimum height=0.1cm, right=1.6cm of sm-bm25-top3-gen] (legend1) {};
    
    \node[anchor=west] at ([xshift=0.1cm]legend1.east) {Retrieval time};

    \node[draw, fill=\gen-color, minimum width=0.1cm, minimum height=0.1cm, below=0.1cm of legend1] (legend2) {};
    \node[anchor=west] at ([xshift=0.1cm]legend2.east) {Generation time};
\end{tikzpicture}
\caption{Response Time Comparison on HotPotQA (Seconds). 
The x-axis represents response time in seconds across different knowledge sizes. 
CAG eliminates retrieval overhead, while dense RAG incurs longer retrieval and generation times due to retrieving and feeding longer text chunks into the LLM. 
Sparse RAG retrieves shorter text spans, resulting in faster generation.
As the knowledge size increases, generation time grows for all methods, but CAG remains competitive while bypassing retrieval completely.
}
\label{fig:time_comparison}
\end{figure}
\autoref{tab:generation-time} and  \autoref{fig:time_comparison} show the retrieval and generation time across different HotPotQA knowledge sizes for various RAG methods and CAG. 
CAG eliminates retrieval time entirely, whereas sparse and dense retrieval-based RAG systems require additional retrieval steps, with dense retrieval incurring higher latency. 
Sparse RAG exhibits insignificant retrieval latency in the experiments, but still requires retrieval before generation. 
As the knowledge size increases, generation time grows across all methods, including CAG, highlighting the computational cost of handling longer contexts. 
However, CAG remains more efficient than dense RAG, as it avoids retrieval overhead while maintaining comparable or superior response times. 
\autoref{tab:generation-time} also compares our CAG approach with standard in-context learning, where the reference text is provided dynamically during inference, requiring real-time KV-cache computation.
The results demonstrate that CAG dramatically reduces generation time, particularly as the reference text length increases. 
This efficiency stems from preloading the KV-cache, which eliminates the need to process the reference text on the fly.

\section{Conclusion}

As long-context LLMs evolve, we present a compelling case for rethinking traditional RAG workflows. 
While our work emphasizes eliminating retrieval latency, there is potential for hybrid approaches that combine preloading with selective retrieval. For example, a system could preload a foundation context and use retrieval only to augment edge cases or highly specific queries. 
This would balance the efficiency of preloading with the flexibility of retrieval, making it suitable for scenarios where context completeness and adaptability are equally important.

\subsection*{Limitations}
Our method requires loading all relevant documents into the models context, making it well-suited for use cases such as internal knowledge bases of small companies, FAQs, and call centers, where the knowledge source is of a manageable size. 
However, this approach becomes impractical for significantly larger datasets. 
Fortunately, as LLMs continue to expand their context lengths and hardware capabilities advance, this limitation is expected to diminish, enabling broader applicability in the future.

\begin{acks}
This work was partially supported by National Science and Technology Council (NSTC), Taiwan, under the grant 112-2221-E-001-016-MY3, by Academia Sinica, under the grant 236d-1120205, and by National Center for High-performance Computing (NCHC), National Applied Research Laboratories (NARLabs), and NSTC under the project ``Trustworthy AI Dialog Engine, TAIDE.'' 
We thank Discover AI\footnote{\url{https://www.youtube.com/watch?v=NaEf_uiFX6o}} and the many individuals who have introduced, shared, and discussed our work, contributing to its broader visibility. 
\end{acks}

\bibliographystyle{ACM-Reference-Format}
\bibliography{sample-sigconf-authordraft}










\end{document}